\documentclass[10pt,letterpaper,twocolumn,british]{article}
\usepackage[T1]{fontenc}
\usepackage[latin9]{inputenc}
\usepackage{array}
\usepackage{multirow}
\usepackage{amsmath}
\usepackage{graphicx}

\makeatletter

\pdfpageheight\paperheight
\pdfpagewidth\paperwidth

\providecommand{\tabularnewline}{\\}

\usepackage{cvpr}
\usepackage{times}
\usepackage{epsfig}
\usepackage{graphicx}

\usepackage{makecell, pict2e}

\usepackage[pagebackref=true,breaklinks=true,letterpaper=true,colorlinks,bookmarks=false]{hyperref}

\cvprfinalcopy 

\def\cvprPaperID{672} 

\ifcvprfinal\fi

\@ifundefined{showcaptionsetup}{}{%
 \PassOptionsToPackage{caption=false}{subfig}}
\usepackage{subfig}
\makeatother

\usepackage{babel}
\begin{document}

\title{Taking a Deeper Look at Pedestrians}

\author{Jan Hosang\qquad{}Mohamed Omran\qquad{}Rodrigo Benenson\qquad{}Bernt
Schiele\vspace{0.5em}
\\
\begin{tabular}{c}
Max Planck Institute for Informatics\tabularnewline
Saarbrücken, Germany\tabularnewline
\texttt{\small{}firstname.lastname@mpi-inf.mpg.de}\tabularnewline
\end{tabular}\vspace{-0.5em}
}
\maketitle
\begin{abstract}
In this paper we study the use of convolutional neural networks (convnets)
for the task of pedestrian detection. Despite their recent diverse
successes, convnets historically underperform compared to other pedestrian
detectors. We deliberately omit explicitly modelling the problem into
the network (e.g.~parts or occlusion modelling) and show that we
can reach competitive performance without bells and whistles. In
a wide range of experiments we analyse small and big convnets, their
architectural choices, parameters, and the influence of different
training data, including pre-training on surrogate tasks.

We present the best convnet detectors on the Caltech and KITTI dataset.
On Caltech our convnets reach top performance both for the Caltech1x
and Caltech10x training setup. Using additional data at training time
our strongest convnet model is competitive even to detectors that
use additional data (optical flow) at test time.
\end{abstract}
\makeatletter 
\renewcommand{\paragraph}{%
\@startsection{paragraph}{4}%
{\z@}{1.0ex \@plus 1ex \@minus .2ex}{-1em}%
{\normalfont \normalsize \bfseries}%
}
\makeatother

\section{\label{sec:Introduction}Introduction}

In recent years the field of computer vision has seen an explosion
of success stories involving convolutional neural networks (convnets).
Such architectures currently provide top results for general object
classification \cite{Krizhevsky2012Nips,Russakovsky2014ArxivImageNet},
general object detection \cite{Szegedy2014Arxiv}, feature matching
\cite{Fischer2014ArxivCnnVersusSift}, stereo matching \cite{Zbontar2014Arxiv},
scene recognition \cite{Zhou2014Nips,Chen2014Arxiv}, pose estimation
\cite{Tompson2014Nips,Chen2014NipsPoseEstimation}, action recognition
\cite{Karpathy2014Cvpr,Simonyan2014Nips} and many other tasks \cite{Razavian2014Arxiv,Azizpour2014Arxiv}.
Pedestrian detection is a canonical case of object detection with
relevant applications in car safety, surveillance, and robotics. A
diverse set of ideas has been explored for this problem \cite{Enzweiler2009PAMI,Geronimo2010Pami,Dollar2011Pami,Benenson2014Eccvw}
and established benchmark datasets are available \cite{Dollar2011Pami,Geiger2012CVPR}.
We would like to know if the success of convnets is transferable to
the pedestrian detection task.

Previous work on neural networks for pedestrian detection has relied
on special-purpose designs, e.g.~hand-crafted features, part and
occlusion modelling. Although these proposed methods perform ably,
current top methods are all based on decision trees learned via Adaboost
\cite{Benenson2014Eccvw,Zhang2014CvprInformedHaar,Paisitkriangkrai2014Eccv,Nam2014Nips,Wang2013IccvRegionlets}.
In this work we revisit the question, and show that both small and
large vanilla convnets can reach top performance on the challenging
Caltech pedestrians dataset. We provide extensive experiments regarding
the details of training, network parameters, and different proposal
methods.
\begin{figure}
\begin{centering}
\hspace*{\fill}\includegraphics[width=1\columnwidth]{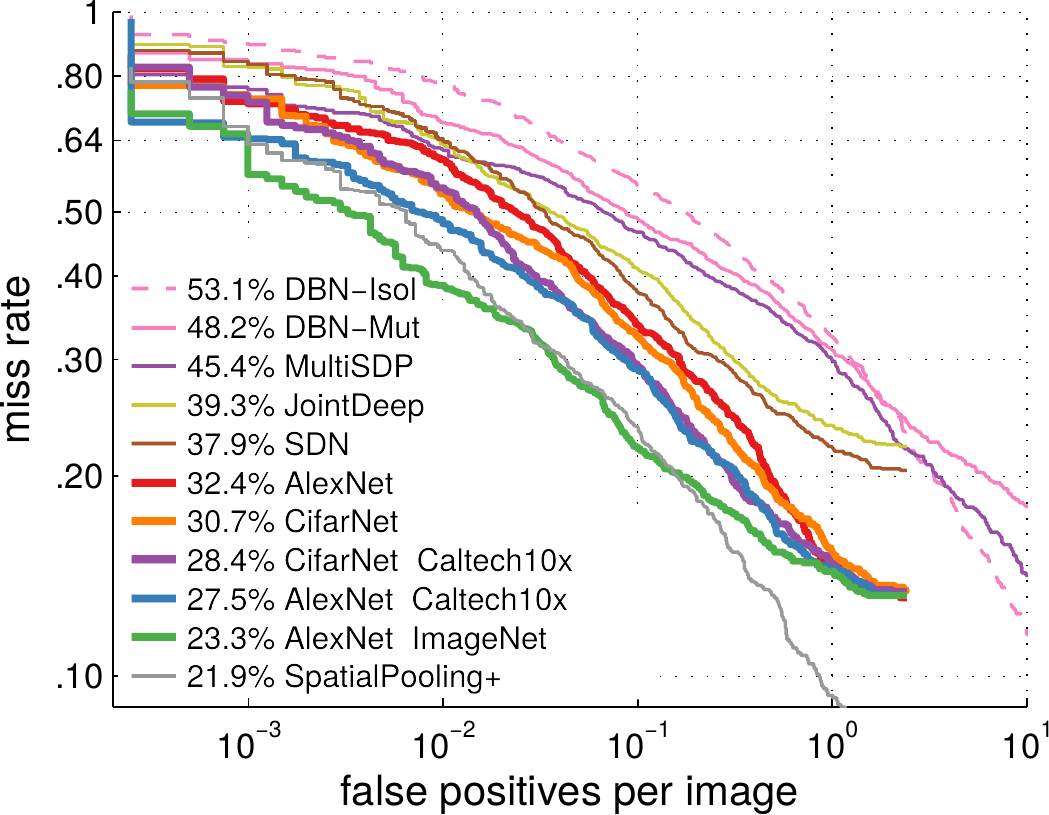}\hspace*{\fill}
\par\end{centering}

\begin{centering}
\vspace{-0em}

\par\end{centering}

\protect\caption{\label{fig:dnn-results}Comparison of convnet methods on the Caltech
test set (see section \ref{sec:Takeaways}). Our CifarNet and AlexNet
results significantly improve over previous convnets, and matches
the best reported results (\texttt{SpatialPooling}+, which additionally
uses optical flow).}
\end{figure}

\subsection{\label{sub:Related-work}Related work}

Despite the popularity of the task of pedestrian detection, only few
works have applied deep neural networks to this task: we are aware
of only six. 

The first paper using convnets for pedestrian detection \cite{Sermanet2013Cvpr}
focuses on how to handle the limited training data (they use the INRIA
dataset, which provides 614 positives and 1218 negative images for
training). First, each layer is initialized using a form of convolutional
sparse coding, and the entire network is subsequently fine-tuned for
the detection task. They propose an architecture that uses features
from the last and second last layer for detection. This method is
named \texttt{Conv\-Net} \cite{Sermanet2013Cvpr}.

A different line of work extends a deformable parts model (\texttt{\small{}DPM})
\cite{Felzenszwalb2010Pami} with a stack of Restricted Boltzmann
Machines (RBMs) trained to reason about parts and occlusion (\texttt{\small{}DBN-Isol})
\cite{Ouyang2012Cvpr}. This model was extended to account for person-to-person
relations (\texttt{\small{}DBN-Mut}) \cite{Ouyang2013Cvpr} and finally
to jointly optimize all these aspects: \texttt{\small{}Joint\-Deep}
\cite{Ouyang2013Iccv} jointly optimizes features, parts deformations,
occlusions, and person-to-person relations. 

The \texttt{\small{}MultiSDP} \cite{Zeng2013Iccv} network feeds each
layer with contextual features computed at different scales around
the candidate pedestrian detection. Finally \texttt{\small{}SDN} \cite{Luo2014Cvpr},
the current best performing convnet for pedestrian detection, uses
additional ``switchable layers'' (RBM variants) to automatically
learn both low-level features and high-level parts (e.g. ``head'',
``legs'', etc.). 

Note that none of the existing papers rely on  a ``straightforward''
convolutional network similar to the original LeNet \cite{LeCun1998Gradient}
(layers of convolutions, non-linearities, pooling, inner products,
and a softmax on top). We will revisit this decision in this paper.

\paragraph{Object detection}

Other than pedestrian detection, related convnets have been used for
detection of ImageNet \cite{Russakovsky2014ArxivImageNet,Krizhevsky2012Nips,He2014Eccv,Szegedy2014Arxiv,Ouyang2014ArxivDeepIdNet,Simonyan2014ArxivVgg}
and Pascal VOC categories \cite{Girshick2014CvprRcnn,Agrawal2014Eccv}.
The most successful general object detectors are based on variants
of the \texttt{\small{}R-CNN} framework \cite{Girshick2014CvprRcnn}.
Given an input image, a reduced set of detection proposals is created,
and these are then evaluated via a convnet. This is essentially a
two-stage cascade sliding window method. See \cite{Hosang2014Bmvc}
for a review of recent proposal methods.

\paragraph{Detection proposals}

The most popular proposal method for generic objects is \texttt{\small{}SelectiveSearch}
\cite{Uijlings2013Ijcv}. The recent review \cite{Hosang2014Bmvc}
also points out \texttt{EdgeBoxes}~\cite{Zitnick2014Eccv} as a fast
and effective method. For pedestrian detection \texttt{\small{}DBN-Isol}
and \texttt{\small{}DBN-Mut} use \texttt{\small{}DPM} \cite{Felzenszwalb2010Pami}
for proposals. \texttt{\small{}JointDeep}, \texttt{\small{}MultiSDP},
and \texttt{\small{}SDN} use a HOG+CSS+linear SVM detector (similar
to \cite{Walk2010Cvpr}) for proposals. Only \texttt{ConvNet} \cite{Sermanet2013Cvpr}
applies a convnet in a sliding fashion.

\paragraph{Decision forests}

Most methods proposed for pedestrian detection do not use convnets
for detection. Leaving aside methods that use optical flow, the current
top performing methods (on Caltech and KITTI datasets) are \texttt{SquaresChnFtrs}
\cite{Benenson2014Eccvw}, \texttt{InformedHaar} \cite{Zhang2014CvprInformedHaar},
\texttt{SpatialPooling} \cite{Paisitkriangkrai2014Eccv}, \texttt{LDCF}
\cite{Nam2014Nips}, and \texttt{Regionlets} \cite{Wang2013IccvRegionlets}.
All of them are boosted decision forests, and can be considered variants
of the integral channels features architecture \cite{Dollar2009Bmvc}.
\texttt{Regionlets} and \texttt{SpatialPooling} use an large set of
features, including HOG, LBP and CSS, while \texttt{SquaresChnFtrs},
\texttt{InformedHaar}, and \texttt{LDCF} build over HOG+LUV. On the
Caltech benchmark, the best convnet (\texttt{\small{}SDN}) is outperformed
by all aforementioned methods.%
\footnote{\texttt{Regionlets} matches \texttt{SpatialPooling} on the KITTI benchmark,
and thus by transitivity would improve over \texttt{\small{}SDN} on
Caltech.%
}

\paragraph{Input to convnets}

It is important to highlight that \texttt{ConvNet} \cite{Sermanet2013Cvpr}
learns to predict from YUV input pixels, whereas all other methods
use additional hand-crafted features. \texttt{DBN-Isol} and \texttt{DBN-Mut}
use HOG features as input. \texttt{MultiSDP} uses HOG+CSS features
as input. \texttt{JointDeep} and \texttt{SDN} uses YUV+Gradients as
input (and HOG+CSS for the detection proposals). We will show in our
experiments that good performance can be reached using RGB alone,
but we also show that more sophisticated inputs systematically improve
detection quality. Our data indicates that the antagonism ``hand-crafted
features versus convnets'' is an illusion.

\subsection{\label{sub:Contributions}Contributions}

In this paper we propose to revisit pedestrian detection with convolutional
neural networks by carefully exploring the design space (number of
layers, filter sizes, etc.), and the critical implementation choices
(training data preprocessing, effect of detections proposal, etc.).
We show that both small ($10^{5}$ parameters) and large ($6\cdot10^{7}$
parameters) networks can reach good performance when trained from
scratch (even when using the same data as previous methods). We also
show the benefits of using extended and external data, which leads
to the strongest single-frame detector on Caltech. We report the best
known performance for a convnet on the challenging Caltech dataset
(improving by more than $10$ percent points), and the first convnet
results on the KITTI dataset.

\section{\label{sec:Training-data}Training data}

It is well known that for convnets the volume of training data is
quite important to reach good performance. Below are the datasets
we consider along the paper.

\paragraph{Caltech}

The Caltech dataset and its associated benchmark~\cite{Dollar2011Pami,Benenson2014Eccvw}
is one of the most popular pedestrian detection datasets. It consists
of videos captured from a car traversing U.S. streets under good weather
conditions. The standard training set in the ``Reasonable'' setting
consists of $4\,250$ frames with $\sim2\cdot10^{3}$ annotated pedestrians,
and the test set covers $4\,024$ frames with $\sim1\cdot10^{3}$
pedestrians.

\paragraph{Caltech validation set}

In our experiments we also use Caltech training data for validation.
For those experiments we use one of the suggested validation splits
\cite{Dollar2011Pami}: the first five training videos are used for
validation training and the sixth training video for validation testing.

\paragraph{Caltech10x}

Because the Caltech dataset videos are fully annotated, the amount
of training data can be increased by resampling the videos. Inspired
by \cite{Nam2014Nips}, we increase the training data tenfold by sampling
one out of three frames (instead of one out of thirty frames in the
standard setup). This yields $\sim2\cdot10^{4}$ annotated pedestrians
for training, extracted from $42\,782$ frames.

\paragraph{KITTI}

The KITTI dataset \cite{Geiger2012CVPR} consists of videos captured
from a car traversing German streets, also under good weather conditions.
Although similar in appearance to Caltech, it has been shown to have
different statistics (see \cite[supplementary material]{Benenson2014Eccvw}).
Its training set contains $4\,445$ pedestrians ($4\,024$ taller
than $40$ pixels) over $7\,481$ frames, and its test set $7\,518$
frames.

\paragraph{ImageNet, Places}

In section \ref{sec:Large-convolutional-network} we will consider
using large convnets that can exploit pre-training for surrogate tasks.
We consider two such tasks (and their associated datasets), the ImageNet
2012 classification of a thousand object categories \cite{Krizhevsky2012Nips,Russakovsky2014ArxivImageNet,Girshick2014CvprRcnn}
and the classification of $205$ scene categories \cite{Zhou2014Nips}.
The datasets provide $1.2\cdot10^{6}$ and $2.5\cdot10^{6}$ annotated
images for training, respectively.

\section{\label{sec:df-to-dnn}From decision forests to neural networks}

Before diving into the experiments, it is worth noting that the proposal
method we are using can be converted into a convnet so that the overall
system can be seen as a cascade of two neural networks.

\texttt{SquaresChnFtrs} \cite{Benenson2013Cvpr,Benenson2014Eccvw}
is a decision forest, where each tree node pools and thresholds information
from one out of several feature channels. As mentioned in section
\ref{sub:Related-work} it is common practice to learn pedestrian
detection convnets on handcrafted features, thus the feature channels
need not be part of the conversion. In this case, a decision node
can be realised using (i) a fully connected layer with constant non-zero
weights corresponding to the original pooling region and zero weights
elsewhere, (ii) a bias term that applies the threshold, (iii) and
a sigmoid non-linearity that yields a decision. A two-layer network
is sufficient to model a level-2 decision tree given the three simulated
node outputs. Finally, the weighted sum over the tree decisions can
be modelled with yet another fully-connected layer.

The mapping from \texttt{SquaresChnFtrs} to a deep neural network
is exact: evaluating the same inputs it will return the exact same
outputs. What is special about the resulting network is that it has
not been trained by back-propagation, but by Adaboost \cite{Bengio2005NipsConvexNeuralNetworks}.
This network already performs better than the best known convnet on
Caltech, \texttt{SDN}. Unfortunately, experiments to soften the non-linearities
and use back-propagation to fine-tune the model parameters did not
show significant improvements.

\section{\label{sec:Vanilla-convolutional-network}Vanilla convolutional networks}

In our experience many convnet architectures and training hyper-parameters
do not enable effective learning for diverse and challenging tasks.
It is thus considered best practice to start exploration from architectures
and parameters that are known to work well and progressively adapt
it to the task at hand. This is the strategy of the following sections.

In this section we first consider CifarNet, a small network designed
to solve the CIFAR-10 classification problem ($10$ objects categories,
$(5+1)\cdot10^{5}$ colour images of $32\negmedspace\times\negmedspace32\ \mbox{pixels}$)
\cite{Krizhevsky2009}. In section \ref{sec:Large-convolutional-network}
we consider AlexNet, a network that has $600$ times more parameters
than CifarNet and designed to solve the ILSVRC2012 classification
problem ($1\,000$ objects categories, $(1.2+0.15)\cdot10^{6}$ colour
images of $\sim$VGA resolution). Both of these networks were introduced
in \cite{Krizhevsky2012Nips} and are re-implemented in the open source
Caffe project~\cite{Jia2014Arxiv}%
\footnote{http://caffe.berkeleyvision.org%
}.

\begin{table*}
\hspace*{\fill}%
\begin{minipage}[t]{0.35\textwidth}%
\begin{center}
\vspace{0em}
\begin{tabular}{cc|c}
Positives & Negatives & MR\tabularnewline
\hline 
\hline 
GT & Random & $83.1\%$\tabularnewline
GT & \texttt{$\mathtt{IoU}<0.5$} & $37.1\%$\tabularnewline
GT & \texttt{$\mathtt{IoU}<0.3$} & $37.2\%$\tabularnewline
GT, \texttt{$\mathtt{IoU}>0.5$} & \texttt{$\mathtt{IoU}<0.5$} & $42.1\%$\tabularnewline
GT, \texttt{$\mathtt{IoU}>0.5$} & \texttt{$\mathtt{IoU}<0.3$} & $41.3\%$\tabularnewline
GT, \texttt{$\mathtt{IoU}>0.75$} & \texttt{$\mathtt{IoU}<0.5$} & $39.9\%$\tabularnewline
\end{tabular}\vspace{-0.5em}

\par\end{center}

\protect\caption{\label{tab:pos-and-negs-definition}Effect of positive and negative
training sets on the detection quality. MR: log-average miss-rate
on Caltech validation set. GT: ground truth bounding boxes.}
\end{minipage}\hspace*{\fill}%
\begin{minipage}[t]{0.2\textwidth}%
\begin{center}
\vspace{0em}
\begin{tabular}{cc}
Window size & MR\tabularnewline
\hline 
\hline 
$32\times32$ & $50.6\%$\tabularnewline
$64\times32$ & $48.2\%$\tabularnewline
$128\times64$ & $39.9\%$\tabularnewline
$128\times128$ & $49.4\%$\tabularnewline
$227\times227$ & $54.9\%$\tabularnewline
\end{tabular}\vspace{-0.5em}

\par\end{center}

\protect\caption{\label{tab:window-size}Effect of the window size on the detection
quality. MR: see table \ref{tab:pos-and-negs-definition}.}
\end{minipage}\hspace*{\fill}%
\begin{minipage}[t]{0.25\textwidth}%
\begin{center}
\vspace{0em}
\begin{tabular}{cc}
Ratio & MR\tabularnewline
\hline 
\hline 
$None$ & $41.4\%$\tabularnewline
$1:10$ & $40.6\%$\tabularnewline
$1:5$ & $39.9\%$\tabularnewline
$1:1$ & $39.8\%$\tabularnewline
\end{tabular}\vspace{-0.5em}

\par\end{center}

\protect\caption{\label{tab:batch-data-ratio}Detection quality as a function of the
strictly enforced ratio of positives:negatives in each training batch.
\textit{None}: no ratio enforced. MR: see table \ref{tab:pos-and-negs-definition}.}
\end{minipage}\hspace*{\fill}
\end{table*}
\begin{figure}
\begin{centering}
\includegraphics[width=1.05\columnwidth]{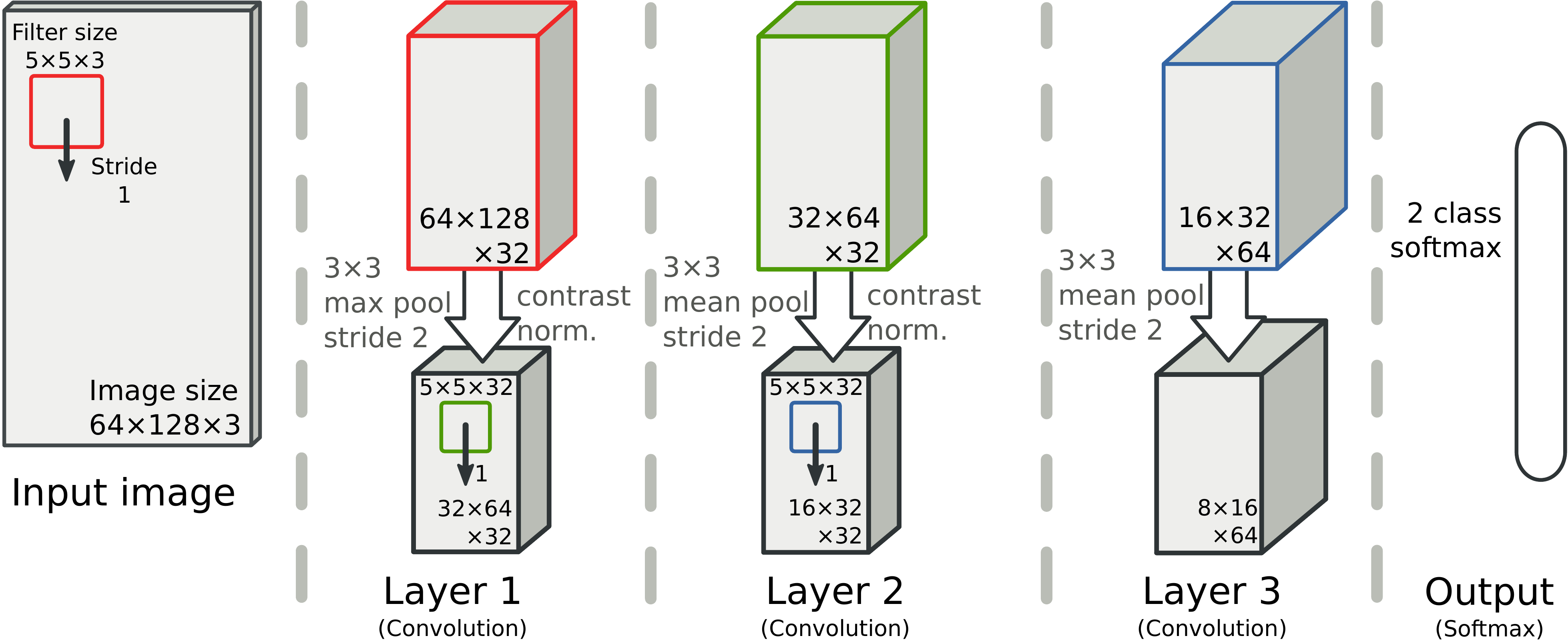}
\par\end{centering}

\protect\caption{\label{fig:CifarNet}Illustration of the CifarNet, $\sim\negthickspace10^{5}$
parameters.}
\end{figure}

Although pedestrian detection is quite a different task than CIFAR-10,
we decide to start our exploration from the CifarNet, which provides
fair performance on CIFAR-10. Its architecture is depicted in figure
\ref{fig:CifarNet}, unless otherwise specified we use raw RGB input.

We first discuss how to use the CifarNet network (section \ref{sub:How-to-use-CifarNet}).
This naive approach already improves over the best known convnets
(section \ref{sub:CifarNetRGB-best-results}). Sections \ref{sub:smallnet-Architectures}
and \ref{sub:smallnet-Input-channels} explore the design space around
CifarNet and further push the detection quality. All models in this
section are trained using Caltech data only (see section~\ref{sec:Training-data}).

\subsection{\label{sub:How-to-use-CifarNet}How to use CifarNet? }

Given an initial network specification, there are still several design
choices that affect the final detection quality. We discuss some of
them in the following paragraphs.

\paragraph{Detection proposals}

\begin{figure}
\begin{centering}
\includegraphics[width=0.85\columnwidth]{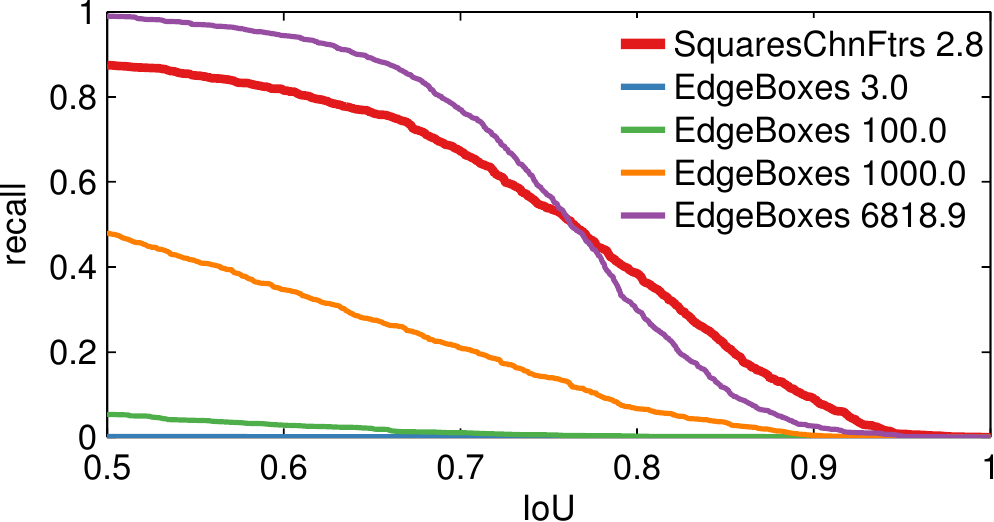}\vspace{0em}

\par\end{centering}

\protect\caption{\label{fig:recall-vs-iou}Recall of ground truth annotations versus
Intersection-over-Union  threshold on the Caltech test set. The legend
indicates the average number of detection proposals per image for
each curve. A pedestrian detector (\texttt{Squa\-res\-ChnFtrs} \cite{Benenson2014Eccvw})
generates much better proposals than a state of the art generic method
(\texttt{Edge\-Boxes}~\cite{Zitnick2014Eccv}). }
\end{figure}

Unless otherwise specified we use the \texttt{\small{}SquaresChnFtrs}
\cite{Benenson2013Cvpr,Benenson2014Eccvw} detector to generate proposals
because, at the time of writing, it is the best performing pedestrian
detector (on Caltech) with source code available. In figure \ref{fig:recall-vs-iou}
we compare \texttt{\small{}SquaresChnFtrs} against \texttt{EdgeBoxes}~\cite{Zitnick2014Eccv},
a state of the art class-agnostic proposal method. Using class-specific
proposals allows to reduce the number of proposals by three orders
of magnitude.

\paragraph{Thresholds for positive and negative samples}

Given both training proposals and ground truth (GT) annotations, we
now consider which training label to assign to each proposal. A proposal
is considered to be a positive example if it exceeds a certain Intersection-over-Union
(IoU) threshold for at least one GT annotation. It is considered negative
if it does not exceed a second IoU threshold for any GT annotation,
and is ignored otherwise. We find that using GT annotations as positives
is beneficial (i.e. not applying significant jitter).

\paragraph{Model window size}

A typical choice for pedestrian detectors is a model window of $128\!\times\!64\ \mbox{pixels}$
in which the pedestrian occupies an area of $96\!\times\!48$ \cite{Dalal2005Cvpr,Dollar2009Bmvc,Benenson2013Cvpr,Benenson2014Eccvw}.
It is unclear that this is the ideal input size for convnets. Despite
CifarNet being designed to operate over $32\!\times\!32\ \mbox{pixels}$,
table~\ref{tab:window-size} shows that a model size of $128\!\times\!64\ \mbox{pixels}$
indeed works best. We experimented with other variants (stretching
versus cropping, larger context border) with no clear improvement.

\paragraph{Training batch}

In a detection setup, training samples are typically highly imbalanced
towards the background class. Although in our validation setup the
imbalance is limited (see table~\ref{tab:batch-data-ratio}), we
found it beneficial throughout our experiments to enforce a strict
ratio of positive to negative examples per batch of the stochastic
gradient descend optimisation. The final performance is not sensitive
to this parameter as long as some ratio (vs. \textit{None}) is maintained.
We use a ratio of $1:5$.

In the supplementary material we detail all other training parameters.

\subsection{\label{sub:CifarNetRGB-best-results}How far can we get with the
CifarNet?}

Given the parameter selection on the validation set from previous
sections, how does CifarNet compare to previously reported convnet
results on the Caltech test set? In table~\ref{tab:cifarnet-proposals}
and figure~\ref{fig:dnn-results}, we see that our naive network
right away improves over the best known convnet ($30.7\%\ \mbox{MR}$
versus \texttt{SDN} $37.9\%\ \mbox{MR}$).

To decouple the contribution of our strong \texttt{\small{}SquaresChnFtrs}
proposals to the CifarNet performance, we also train a CifarNet using
the proposal from \texttt{\small{}JointDeep}~\cite{Ouyang2013Iccv}.
When using the same detection proposals at training and test time,
the vanilla CifarNet already improves over both custom-designed \texttt{\small{}JointDeep}
and \texttt{SDN}.

Our CifarNet results are surprisingly close to the best known pedestrian
detector trained on Caltech1x ($30.7\%\ \mbox{MR}$ versus \texttt{Spatial\-Pooling}
$29.2\%\ \mbox{MR}$ \cite{Paisitkriangkrai2014Eccv}).

\subsection{\label{sub:smallnet-Architectures}Exploring different architectures}

Encouraged by our initial results, we proceed to explore different
parameters for the CifarNet architecture.

\subsubsection{\label{sub:Number-and-size-of-filters}Number and size of convolutional
filters}

Using the Caltech validation set we perform a swipe of convolutional
filter sizes ($3\!\times\!3$, $5\!\times\!5$, or $7\!\times\!7$
pixels) and number of filters at each layer ($16$, $32$, or $64$
filters). We include the full table in the supplementary material.
We observe that using large filter sizes hurts quality, while the
varying the number of filters shows less impact. Although some fluctuation
in miss-rate is observed, overall there is no clear trend indicating
that a configuration is clearly better than another. Thus, for sake
of simplicity, we keep using CifarNet ($32$-$32$-$64$ filters of
$5\!\times\!5$ pixel) in the subsequent experiments.

\subsubsection{\label{sub:Number-and-type-of-layers}Number and type of layers}

In table \ref{tab:smallnet-architectures-quality} we evaluate the
effect of changing the number and type of layers, while keeping other
CifarNet parameters fix. Besides convolutional layers (CONV) and fully-connected
layers (FC), we also consider locally-connected layers (LC) \cite{Taigman2014Cvpr},
and concatenating features across layers (CONCAT23) (used in \texttt{Conv\-Net}
\cite{Sermanet2013Cvpr}). None of the considered architecture changes
improves over the original three convolutional layers of CifarNet.

\begin{table}
\begin{centering}
\begin{tabular}{cc|l}
Method & Proposal & Test MR\tabularnewline
\hline 
\hline 
{\small{}Proposals of }\cite{Ouyang2013Iccv} & \texttt{\small{}-} & $45.5\%$ \tabularnewline
\texttt{\small{}JointDeep} & {\small{}Proposals of }\cite{Ouyang2013Iccv} & $39.3\%$ \cite{Ouyang2013Iccv}\tabularnewline
\texttt{\small{}SDN} & {\small{}Proposals of }\cite{Ouyang2013Iccv} & $37.9\%$ \cite{Luo2014Cvpr}\tabularnewline
CifarNet & {\small{}Proposals of }\cite{Ouyang2013Iccv} & $36.5\%$\tabularnewline
\hline 
\texttt{\small{}SquaresChnFtrs} & - & $34.8\%$ \cite{Benenson2014Eccvw}\tabularnewline
CifarNet & \texttt{\small{}SquaresChnFtrs} & $\mathit{30.7\%}$\tabularnewline
\end{tabular}
\par\end{centering}

\protect\caption{\label{tab:cifarnet-proposals}Detection quality as a function of
the method and the proposals used for training and testing (MR: log-average
miss-rate on Caltech test set). When using the exact same training
data as \texttt{JointDeep}, our vanilla CifarNet already improves
over the previous best known convnet on Caltech (\texttt{\small{}SDN}).}
\end{table}
\begin{table}
\begin{centering}
\begin{tabular}{clc}
{\footnotesize{}\#} & \multirow{2}{*}{Architecture} & \multirow{2}{*}{MR}\tabularnewline
{\footnotesize{}layers} &  & \tabularnewline
\hline 
\hline 
\multirow{3}{*}{3} & {\small{}CONV1 CONV2 CONV3} {\footnotesize{}(CifarNet, fig. \ref{fig:CifarNet})} & $\mathit{37.1\%}$\tabularnewline
 & {\small{}CONV1 CONV2 LC} & $43.2\%$\tabularnewline
 & {\small{}CONV1 CONV2 FC} & $47.6\%$\tabularnewline
\hline 
\multirow{4}{*}{4} & {\small{}CONV1 CONV2 CONV3 FC} & $39.6\%$\tabularnewline
 & {\small{}CONV1 CONV2 CONV3 LC} & $40.5\%$\tabularnewline
 & {\small{}CONV1 CONV2 FC1 FC2} & $43.2\%$\tabularnewline
 & {\small{}CONV1 CONV2 CONV3 CONV4} & $43.3\%$\tabularnewline
\hline 
{\small{}DAG} & {\small{}CONV1 CONV2 CONV3 CONCAT23 FC} & $38.4\%$\tabularnewline
\end{tabular}
\par\end{centering}

\protect\caption{\label{tab:smallnet-architectures-quality}Detection quality of different
network architectures (MR: log-average miss-rate on Caltech validation
set), sorted by number of layers before soft-max. DAG: directed acyclic
graph.}
\end{table}

\subsection{\label{sub:smallnet-Input-channels}Input channels}

As discussed in section \ref{sub:Related-work}, the majority of previous
convnets for pedestrian detection use gradient and colour features
as input, instead of raw RGB. In table~\ref{tab:input-channels}
we evaluate the effect of different input features over CifarNet.
It seems that HOG+L channel provide a small advantage over RGB. 

For purposes of direct comparison with the large networks, in the
next sections we keep using raw RGB as input for our CifarNet experiments.
We report the CifarNet test set results in section \ref{sec:Small-or-big}.

\begin{table}
\begin{centering}
\begin{tabular}{cc|c}
Input channels & \# channels & CifarNet\tabularnewline
\hline 
\hline 
RGB & 3 & $39.9\%$\tabularnewline
LUV & 3 & $46.5\%$\tabularnewline
G+LUV & 4 & $40.0\%$\tabularnewline
HOG+L & 7 & $36.8\%$\tabularnewline
HOG+LUV & 10 & $40.7\%$\tabularnewline
\end{tabular}
\par\end{centering}

\protect\caption{\label{tab:input-channels}Detection quality when changing the input
channels network architectures. Results in MR; log-average miss-rate
on Caltech validation set. G indicates luminance channel gradient,
HOG indicates G plus G spread over six orientation bins (hard-binning).
These are the same input channels used by our \texttt{\small{}Squa\-res\-Chn\-Ftrs}
proposal method.}
\end{table}

\section{\label{sec:Large-convolutional-network}Large convolutional network}

One appealing characteristic of convnets is their ability to scale
in size of training data volume. In this section we explore larger
networks trained with more data.

We base our experiments on the \texttt{R-CNN }\cite{Girshick2014CvprRcnn}
approach, which is currently one of the best performer on the Pascal
VOC detection task \cite{Everingham2014Ijcv}. We are thus curious
to evaluate its performance for pedestrian detection.

\subsection{\label{sub:Surrogate-tasks}Surrogate tasks for improved detections}

\begin{figure*}
\begin{centering}
\includegraphics[height=11em]{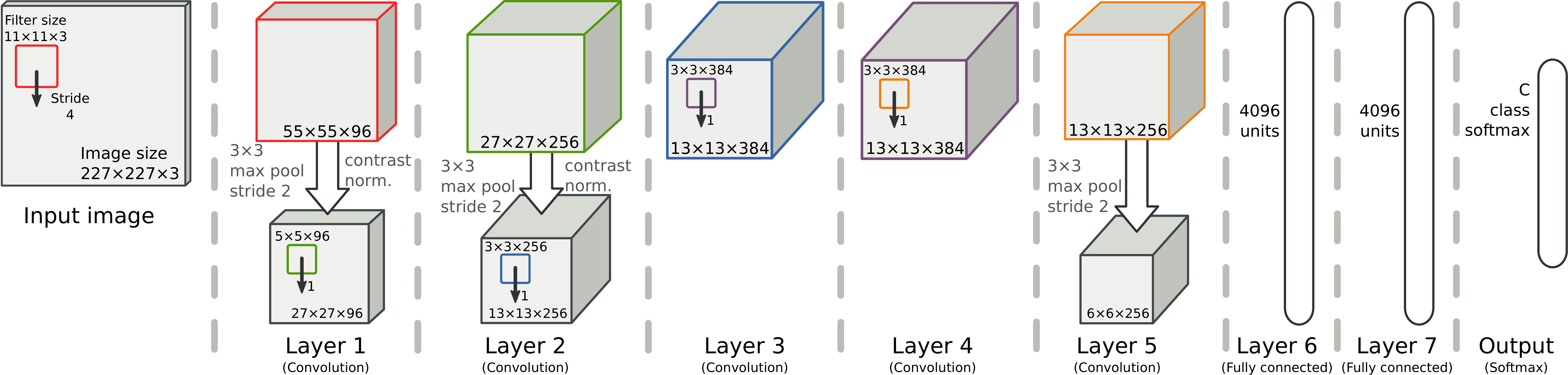}
\par\end{centering}

\protect\caption{\label{fig:AlexNet}Illustration of the AlexNet architecture, $\sim\negthickspace6\cdot10^{7}$
parameters.}
\end{figure*}

The \texttt{R-CNN} approach \texttt{(}``Regions with CNN features'')
wraps the large network previously trained for the ImageNet classification
task \cite{Krizhevsky2012Nips}, which we refer to as AlexNet (see
figure \ref{fig:AlexNet}). We also use ``AlexNet'' as shorthand
for ``\texttt{R-CNN} with AlexNet'' with the distinction made clear
by the context. During \texttt{R-CNN} training AlexNet is fine-tuned
for the (pedestrian) detection task, and in a second step, the softmax
output is replaced by a linear SVM. Unless otherwise specified, we
use the default parameters of the open source, Caffe based, \texttt{R-CNN}
implementation%
\footnote{https://github.com/rbgirshick/rcnn%
}. Like in the previous sections, we use \texttt{\small{}SquaresChnFtrs}
for detection proposals. 

\paragraph{Pre-training}

If we only train the top layer SVM, without fine-tuning the lower
layers of AlexNet, we obtain $39.8\%\ \mbox{MR}$ on the Caltech test
set. This is already surprisingly close to the best known convnet
for the task (\texttt{SDN} $37.9\%\ \mbox{MR}$). When fine-tuning
all layers on Caltech, the test set performance increases dramatically,
reaching $25.9\%\ \mbox{MR}$. This confirms the effectiveness of
the general \texttt{R-CNN} recipe for detection (train AlexNet on
ImageNet, fine-tune for the task of interest).\\
In table \ref{tab:AlexNet-finetuning-results} we investigate the
influence of the pre-training task by considering AlexNets that have
been trained for scene recognition \cite{Zhou2014Nips} (``Places'',
see section \ref{sec:Training-data}) and on both Places and ImageNet
(``Hybrid''). ``Places'' provides results close to ImageNet, suggesting
that the exact pre-training task is not critical and that there is
nothing special about ImageNet.

\paragraph{Caltech10x}

Due to the large number of parameters of AlexNet, we consider providing
additional training data using Caltech10x for fine-tuning the network
(see section~\ref{sec:Training-data}). Despite the strong correlation
across training samples, we do observe further improvement (see table
\ref{tab:AlexNet-finetuning-results}). Interestingly, the bulk of
the improvement is due to more pedestrians (Positives10x, uses positives
from Caltech10x and negatives from Caltech1x). Our top result, $23.3\%\ \mbox{MR}$,
makes our AlexNet setup the best reported single-frame detector on
Caltech (i.e.~no optical flow).

\begin{table}
\begin{centering}
\begin{tabular}{ccc|c}
AlexNet & Fine- & SVM  & \multirow{2}{*}{Test MR}\tabularnewline
training & tuning & training & \tabularnewline
\hline 
\hline 
Random & none & Caltech1x & $86.7\%$\tabularnewline
ImageNet & none & Caltech1x & $39.8\%$\tabularnewline
\hline 
{\small{}P+Imagenet} & \multirow{3}{*}{Caltech1x} & \multirow{3}{*}{Caltech1x} & $30.1\%$\tabularnewline
P: Places &  &  & $27.0\%$\tabularnewline
ImageNet &  &  & $25.9\%$\tabularnewline
\hline 
\multirow{2}{*}{ImageNet} & Positives10x & Positives10x & $23.8\%$\tabularnewline
 & Caltech10x & Caltech10x & $\mathit{23.3\%}$\tabularnewline
\hline 
\multirow{2}{*}{Caltech1x} & - & Caltech1x & $32.4\%$\tabularnewline
 & - & Caltech10x & $32.2\%$ \tabularnewline
\hline 
\multirow{2}{*}{Caltech10x} & - & Caltech1x & $\mathit{27.4\%}$\tabularnewline
 & - & Caltech10x & $\mathit{27.5\%}$\tabularnewline
\hline 
\hline 
\multicolumn{3}{c|}{\texttt{\small{}SquaresChnFtrs} \cite{Benenson2014Eccvw}} & $34.8\%$\tabularnewline
\end{tabular}
\par\end{centering}

\protect\caption{\label{tab:AlexNet-master-table}\label{tab:AlexNet-finetuning-results}\label{tab:AlexNet-no-finetunning-results}Detection
quality when using different training data in different training stages
of the AlexNet: initial training of the convnet, optional fine-tuning
of the convnet, and the SVM training. Positives10x: positives from
Caltech10x and negatives from Caltech1x. Detection proposals provided
by \texttt{\small{}SquaresChnFtrs}, result included for comparison.
See section \ref{sub:Surrogate-tasks} and \ref{sub:AlexNet-Caltech-only}
for details.}
\end{table}

\subsection{\label{sub:AlexNet-Caltech-only}Caltech-only training}

To compare with CifarNet, and to verify whether pre-training is necessary
at all, we train AlexNet ``from scratch'' using solely the Caltech
training data. We collect results in table~\ref{tab:AlexNet-no-finetunning-results}.

Training AlexNet solely on Caltech, yields $32.4\%\ \mbox{MR}$, which
improves over the proposals (\texttt{SquaresChnFtrs} $34.8\%\ \mbox{MR}$)
and the previous best known convnet on Caltech (\texttt{SDN} $39.8\%\ \mbox{MR}$).
Using Caltech10x further improves the performance, down to $27.5\%\ \mbox{MR}$. 

Although these numbers are inferior than the ones obtained with ImageNet
pre-training ($23.3\%\ \mbox{MR}$, see table \ref{tab:AlexNet-finetuning-results}),
we can get surprisingly competitive results using only pedestrian
data despite the $10^{7}$ free parameters of the AlexNet model. AlexNet
with Caltech10x is second best known single-frame pedestrian detector
on Caltech (best known is \texttt{LDCF} $24.8\%\ \mbox{MR}$, which
also uses Caltech10x).

\subsection{\label{sub:AlexNet-Additional-experiments}Additional experiments}

\paragraph{How many layers?}

So far all experiments use the default parameters of \texttt{R-CNN}.
Previous works have reported that, depending on the task, using features
from lower AlexNet layers can provide better results \cite{Agrawal2014Eccv,Razavian2014Arxiv,Azizpour2014Arxiv}.
Table~\ref{tab:AlexNet-which-layer} reports Caltech validation results
when training the SVM output layer on top of layers four to seven
(see figure \ref{fig:AlexNet}). We report results when using the
default parameters and parameters that have been optimised by grid
search (detailed grid search included in supplementary material).\\
We observe a negligible difference between default and optimized parameter
(at most $1$ percent points). Results for default parameters exhibit
a slight trend of better performance for higher levels. These validation
set results indicate that, for pedestrian detection, the \texttt{R-CNN}
default parameters are a good choice overall.

\begin{table}
\begin{centering}
\begin{tabular}{c|cccc}
Parameters & fc7 & fc6 & pool5 & conv4\tabularnewline
\hline 
\hline 
Default & $32.2\%$ & $32.5\%$ & $33.4\%$ & $42.7\%$\tabularnewline
\hline 
Best & $32.0\%$ & $31.8\%$ & $32.5\%$ & $42.4\%$\tabularnewline
\end{tabular}
\par\end{centering}

\protect\caption{\label{tab:AlexNet-which-layer}Detection quality when training the
R-CNN SVM over different layers of the finetuned CNN. Results in MR;
log-average miss-rate on Caltech validation set. ``Best parameters''
are found by exhaustive search on the validation set.}
\end{table}

\paragraph{Effect of proposal method}

When comparing the performance of AlexNet fine-tuned on Caltech1x
to the proposal method, we see an improvement of $9\ \mbox{pp}$ (percent
points) in miss-rate. In table~\ref{tab:AlexNet-effect-of-proposals}
we study the impact of using weaker or stronger proposals. Both \texttt{ACF}
\cite{Dollar2014Pami} and \texttt{SquaresChnFtrs} \cite{Benenson2013Cvpr,Benenson2014Eccvw}
provide source code, allowing us to generate training proposals. \texttt{Katamari}
\cite{Benenson2014Eccvw} and \texttt{SpatialPooling+ }{\small{}\cite{Paisitkriangkrai2014Eccv}}
are current top performers on the Caltech dataset, both using optical
flow, i.e.~additional information at test time. There is a $\sim\negthickspace10\ \mbox{pp}$
gap between the detectors \texttt{ACF}, \texttt{SquaresChnFtrs}, and
\texttt{Katamari}/\texttt{SpatialPooling}, allowing us to cover different
operating points.\\
The results of table~\ref{tab:AlexNet-effect-of-proposals} indicate
that, despite the $10\ \mbox{pp}$ gap, there is no noticeable difference
between AlexNet models trained with \texttt{ACF} or \texttt{SquaresChnFtrs}.
It is seems that as long as the proposals are not random (see top
row of table \ref{tab:pos-and-negs-definition}), the obtained quality
is rather stable. The results also indicate that the quality improvement
from AlexNet saturates around $\sim\!22\%\ \mbox{MR}$. Using stronger
proposals does not lead to further improvement. This means that the
discriminative power of our trained AlexNet is on par with the best
known models on the Caltech dataset, but does not overtake them.

\begin{table}
\begin{centering}
\begin{tabular}{ccc|cl}
\multirow{1}{*}{Fine-} & Training & Testing & \multirow{2}{*}{Test MR} & $\Delta$ vs.\tabularnewline
tuning & proposals & proposals &  & proposals\tabularnewline
\hline 
\hline 
\multirow{6}{*}{$1\times$} & \texttt{\small{}ACF} & \texttt{\small{}ACF} & $34.5\%$ & $9.7\%$\tabularnewline
 & \texttt{\small{}SCF} & \texttt{\small{}ACF} & $34.3\%$ & $9.9\%$\tabularnewline
\cline{2-5} 
 & \texttt{\small{}ACF} & \texttt{\small{}SCF} & $26.9\%$ & $7.9\%$\tabularnewline
 & \texttt{\small{}SCF} & \texttt{\small{}SCF} & $25.9\%$ & $8.9\%$\tabularnewline
\cline{2-5} 
 & \texttt{\small{}ACF} & \texttt{\small{}Katamari} & $25.1\%$ & $-2.6\%$\tabularnewline
 & \texttt{\small{}SCF} & \texttt{\small{}Katamari} & $24.2\%$ & $-1.7\%$\tabularnewline
\hline 
\hline 
\multirow{4}{*}{$10\times$} & \texttt{\small{}SCF} & \texttt{\small{}LDCF} & $23.4\%$ & $1.4\%$\tabularnewline
 & \texttt{\small{}SCF} & \texttt{\small{}SCF} & $23.3\%$ & $11.5\%$\tabularnewline
 & \texttt{\small{}SCF} & \texttt{\small{}SP+} & $22.0\%$ & $-0.1\%$\tabularnewline
 & \texttt{\small{}SCF} & \texttt{\small{}Katamari} & $21.6\%$ & $0.9\%$\tabularnewline
\hline 
\hline 
\multicolumn{3}{c|}{\texttt{\small{}ACF}{\small{} \cite{Dollar2014Pami}}} & $44.2\%$ & \tabularnewline
\multicolumn{3}{c|}{\texttt{\small{}SCF}: \texttt{\small{}SquaresChnFtrs}{\small{} \cite{Benenson2014Eccvw}}} & $34.8\%$ & \tabularnewline
\multicolumn{3}{c|}{\texttt{\small{}LDCF}{\small{} \cite{Nam2014Nips}}} & $24.8\%$ & \tabularnewline
\multicolumn{3}{c|}{\texttt{\small{}Katamari}{\small{} \cite{Benenson2014Eccvw}}} & $22.5\%$ & \tabularnewline
\multicolumn{3}{c|}{\texttt{\small{}SP+}: \texttt{\small{}SpatialPooling+}{\small{} \cite{Paisitkriangkrai2014Arxiv}}} & $21.9\%$ & \tabularnewline
\end{tabular}
\par\end{centering}

\protect\caption{\label{tab:AlexNet-effect-of-proposals}Effect of proposal methods
on detection quality of \texttt{R-CNN}. $1\times$/$10\times$ indicates
fine-tuning on Caltech or Caltech10x. Test MR: log-average miss rate
on Caltech test set. $\Delta$: the improvement in MR of the rescored
proposals over the test proposals alone.\texttt{\small{}}}
\end{table}

\paragraph{KITTI test set}

\begin{figure}
\begin{centering}
\includegraphics[width=1\columnwidth]{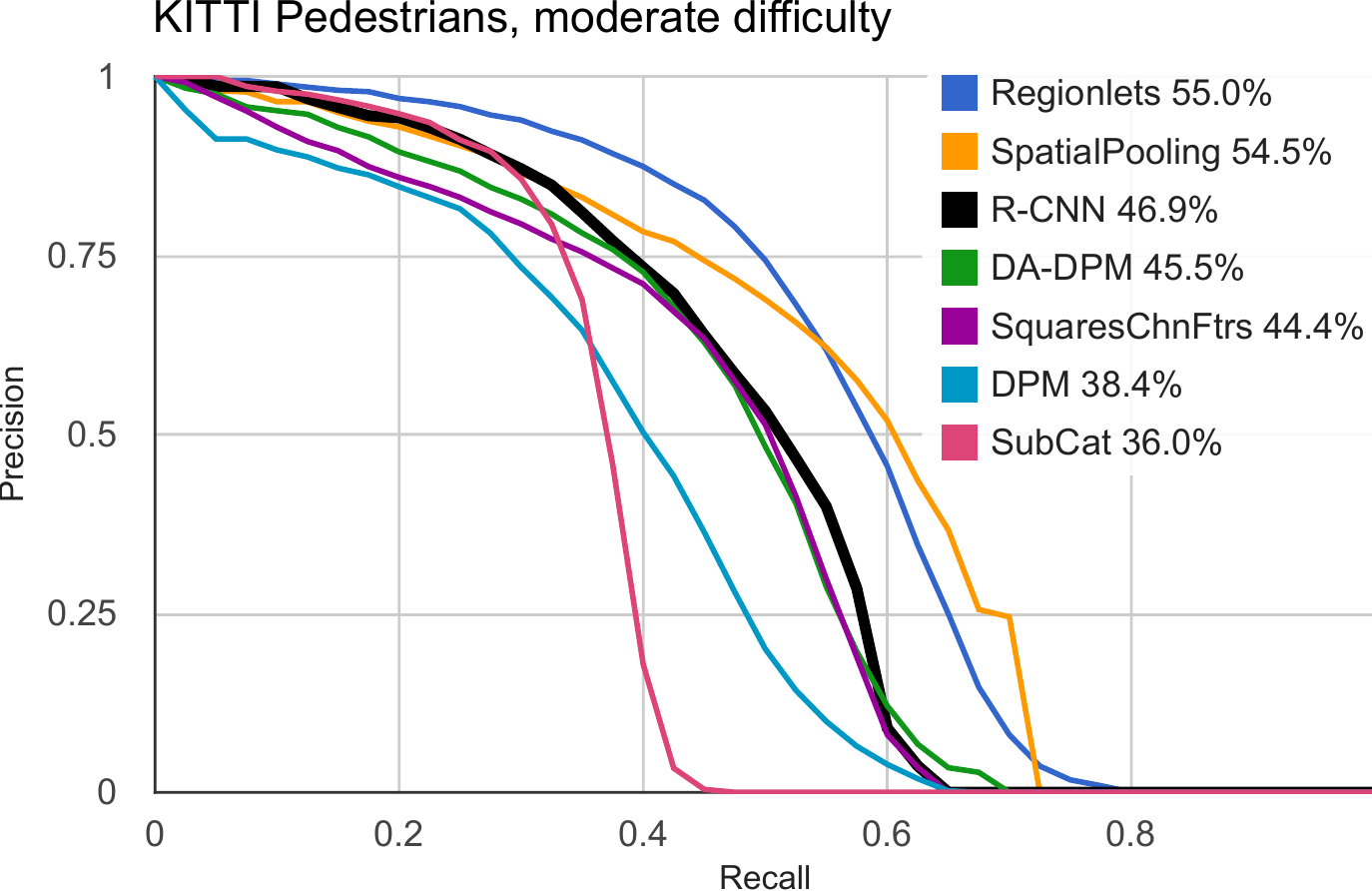}
\par\end{centering}

\begin{centering}
\vspace{-0em}

\par\end{centering}

\protect\caption{\label{fig:kitti-results}AlexNet over on KITTI test set.}
\end{figure}

In figure~\ref{fig:kitti-results} we show performance of the AlexNet
in context of the KITTI pedestrian detection benchmark~\cite{Geiger2012CVPR}.
The network is pre-trained on ImageNet and fine-tuned using KITTI
training data. \texttt{SquaresChnFtrs} reaches $44.4\%\ \mbox{AP}$
(average precision), which the AlexNet can improve to $46.9\%\ \mbox{AP}$.
These are the first published results for convnets on the KITTI pedestrian
detection dataset.

\subsection{\label{sub:AlexNet-Failures}Error analysis}

Results from the previous section are encouraging, but not as good
as could be expected from looking at improvements on Pascal VOC. So
what bounds performance? The proposal method? The localization quality
of the convnet?

Looking at the highest scoring false positives paints a picture of
localization errors of the proposal method, the \texttt{R-CNN}, and
even the ground truth. To quantify this effect we rerun the Caltech
evaluation but remove all false positives that touch an annotation.
This experiment provides an upper bound on performance when solving
localisation issues in detectors and doing perfect non-maximum suppression.
We see a surprisingly consistent improvement for all methods of about
$2\%$ MR. This means that the intuition we gathered from looking
at false positives is wrong and actually almost all of the mistakes
that worsen the MR are actually background windows that are mistaken
for pedestrians. What is striking about this result is that this is
not just the case for our R-CNN experiments on detection proposals
but also for methods that are trained as a sliding window detector.

\section{\label{sec:Small-or-big}Small or big convnet?}

\begin{table}
\begin{centering}
\begin{tabular}{cc|cc}
\multirow{1}{*}{Architecture} & \# & \multicolumn{2}{c}{Test MR}\tabularnewline
training & parameters & Caltech1x & Caltech10x\tabularnewline
\hline 
\hline 
CifarNet & $\sim\negthinspace10^{5}$ & $30.7\%$ & $28.4\%$\tabularnewline
MediumNet & $\sim\negthinspace10^{6}$ & $-$ & $27.9\%$\tabularnewline
AlexNet & $\sim\negthinspace10^{7}$ & $32.4\%$ & $27.5\%$\tabularnewline
\hline 
\hline 
\multicolumn{2}{c|}{\texttt{SquaresChnFtrs} \cite{Benenson2014Eccvw}} & $34.8\%$ & \tabularnewline
\end{tabular}
\par\end{centering}

\protect\caption{\label{tab:small-versus-big}Selection of results (presented in previous
sections) when training different networks using Caltech training
data only. MR: log-average miss-rate on Caltech test set. See section
\ref{sec:Small-or-big}.}
\end{table}
Since we have analysed the CifarNet and AlexNet separately, we compare
their performance in this section side by side. Table~\ref{tab:small-versus-big}
shows performance on the Caltech test set for models that have been
trained only on Caltech1x and Caltech10x. With less training data
the CifarNet reaches $30.7\%$ MR, performing 2 percent points better
than the AlexNet. On Caltech10x, we find the CifarNet performance
improved to $28.4\%$, while the AlexNet improves to $27.1\%$ MR.
The trend confirms the intuition that models with lower capacity saturate
earlier when increasing the amount of training data than models with
higher capacity. We can also conclude that the AlexNet would profit
from better regularisation when training on Caltech1x.

\paragraph{Timing}

The runtime during detection is about 3ms per proposal window. This
is too slow for sliding window detection, but given a fast proposal
method that has high recall with less than $100$ windows per image,
scoring takes about 300ms per image. In our experience \texttt{SquaresChnFtrs}
runs in 2s per image, so proposing detections takes most of the detection
time.

\section{\label{sec:Takeaways}Takeaways}

\begin{figure}
\begin{centering}
\hspace*{\fill}\includegraphics[width=1\columnwidth]{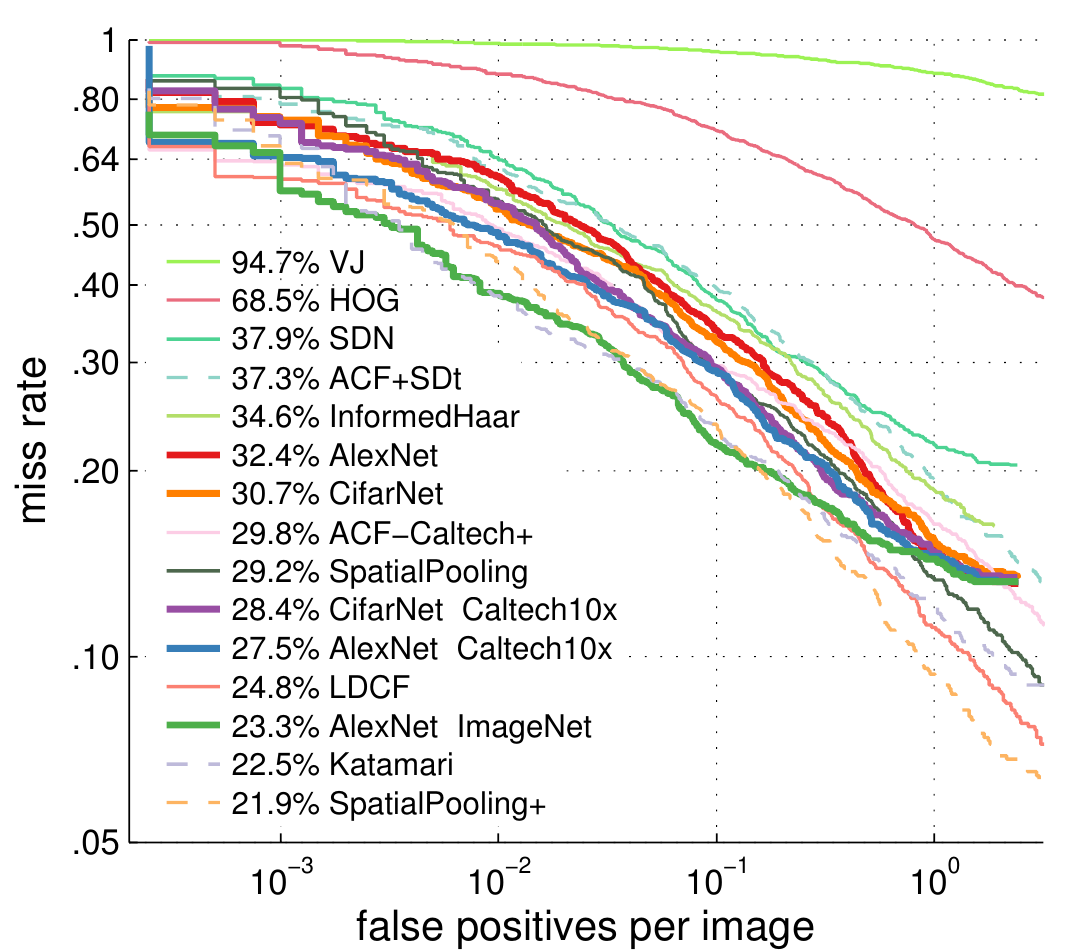}\hspace*{\fill}
\par\end{centering}

\begin{centering}
\vspace{-0em}

\par\end{centering}

\protect\caption{\label{fig:caltech-results}Comparison of our key results (thick lines)
with published methods on Caltech test set. Methods using optical
flow are dashed.}
\end{figure}
 Previous work suggests that convnets for pedestrian detection underperform,
despite having involved architectures (see~\cite{Benenson2014Eccvw}
for a survey of pedestrian detection). In this paper we showed that
neither has to be the case. We present a wide range of experiments
with two off-the-shelf models that reach competitive performance:
the small CifarNet and the big AlexNet.

We present two networks that are trained on Caltech only, which outperform
all previously published convnets on Caltech. The CifarNet shows better
performance than related work, even when using the same training data
as the respective methods (section~\ref{sub:CifarNetRGB-best-results}).
Despite its size, the AlexNet also improves over all convnets even
when it is trained on Caltech only (section~\ref{sub:AlexNet-Caltech-only}).

We push the state of the art for pedestrian detectors that have been
trained on Caltech1x and Caltech10x. The CifarNet is the best single-frame
pedestrian detector that has been trained on Caltech1x (section~\ref{sub:CifarNetRGB-best-results}),
while AlexNet is the best single-frame pedestrian detector trained
on Caltech10x (section~\ref{sub:AlexNet-Caltech-only}).

In figure~\ref{fig:caltech-results}, we include include all published
methods on Caltech into the comparison, which also adds methods that
use additional information at test time. The AlexNet that has been
pre-trained on ImageNet reaches competitive results to the best published
methods, but without using additional information at test time (section~\ref{sub:Surrogate-tasks}).

We report first results for convnets on the KITTI pedestrian detection
benchmark. The AlexNet improves over the proposal method alone, delivering
encouraging results to further push KITTI performance with convnets.

\section{\label{sec:Conclusion}Conclusion}

We have presented extensive and systematic experimental evidence
on the effectiveness of convnets for pedestrian detection. Compared
to previous convnets applied to pedestrian detection our approach
avoids custom designs. When using the exact same proposals and training
data as previous approaches our ``vanilla'' networks outperform
previous results.

We have shown that with pre-training on surrogate tasks, convnets
can reach top performance on this task. Interestingly we have shown
that even without pre-training competitive results can be achieved,
and this result is quite insensitive to the model size (from $10^{5}$
to $10^{7}$ parameters). Our experiments also detail which parameters
are most critical to achieve top performance. We report the best
known results for convnets on both the challenging Caltech and KITTI
datasets.

Our experience with convnets indicate that they show good promise
on pedestrian detection, and that reported best practices do transfer
to said task. That being said, on this more mature field we do not
observe the large improvement seen on datasets such as Pascal VOC
and ImageNet.

\paragraph{Acknowledgements}

We thank Shanshan Zhang for the help provided setting up some of the
experiments. \bibliographystyle{ieee}
\bibliography{2015_cvpr_dnn_for_pedestrian_detection}
\clearpage{}

\appendix

\section{CifarNet training, the devil is in the details}

Training neural networks is sensitive to a large number of parameters,
including the learning rate, how the network weights are initialised,
the type of regularisation applied to the weights, and the training
batch size. It is difficult to isolate the effects of the individual
parameters, and the best parameters will largely depend on the specific
setup. We report here the parameters we used.

We train CifarNet via stochastic gradient descent (SGD) with a learning
rate of $0.005$, a momentum of $0.9$, and a batch size of $128$.
After $60$ epochs, we reduce the learning rate by a factor of $0.1$
and train for an additional $10$ epochs. Reducing the learning rate
even further did not improve the classification accuracy. The other
learning rate policies we explored yielded inferior performance (e.g.
gradually reducing the learning rate each training iteration). Careful
tuning of the learning rate while adjusting the batch size was critical.

Other than the softmax classification loss, the training loss includes
a $\mbox{L}2$ regularisation of the network weights. In the objective
function, this regularization term has a weight of $0.005$ for all
layers but the last one (softmax weights), which receives weight $1$.
This parameter is referred in \texttt{Caffe} as ``weight decay''. 

The network weights are initialised by drawing values from a Gaussian
distribution with standard deviation $\sigma=0.01$, with the exception
of the first layer, for which we set $\sigma=0.0001$.

\section{\label{sec:CifarNet-grid-search}Grid search around CifarNet}

Table \ref{tab:num-filters-vs-filters-size} shows the detection quality
of different variants of CifarNet obtained by changing the number
and size of the convolutional filters of each layer. See related section
4.3.1 of the main paper. Since different training rounds have different
random initial weights, we train four networks for each parameter
set and average the results. We report both mean and standard deviation
of the miss rate on our validation set.

We observe that using either too small or too large filter sizes throughout
the network hurts quality.  The network width also seems to matter,
a network too narrow or too wide can negatively impact classification
accuracy. All and all the ``middle-section'' of the table shows
only small fluctuations in miss-rate (specially when considering the
variance).

In addition to filter size and layer width, we also experimented with
different types of pooling layers (max-pooling versus mean-pooling),
see figure 2 of main paper. Other than on the first layer, replacing
mean-pooling with max-pooling hurts performance. 

The results of table $2$ indicate that there is no set of parameters
close to CifarNet with a clear advantage over the default CifarNet
parameters. When going too far from CifarNet parameters, classification
accuracy plunges.

\begin{table*}
\begin{centering}
\begin{tabular}{c|ccccccc|c}
\diaghead{\hskip4.4em}{Sizes}{\# filters} & {\footnotesize{}$16,16,16$} & {\footnotesize{}$\mathit{32,32,64}$} & {\footnotesize{}$32,64,32$} & {\footnotesize{}$64,32,32$} & {\footnotesize{}$32,32,32$} & {\footnotesize{}$64,64,64$} & {\footnotesize{}$64,32,16$} & Mean\tabularnewline
\hline 
\hline 
{\small{}$3,3,3$} & $48.4\pm1.7$ & $44.4\pm1.0$ & $43.6\pm0.8$ & $45.1\pm1.1$ & $45.2\pm0.7$ & $42.3\pm1.3$ & $46.6\pm2.1$ & $45.1$\tabularnewline
{\small{}$\mathit{5,5,5}$} & $42.7\pm4.2$ & $41.1\pm1.3$ & $39.1\pm1.0$ & $38.9\pm1.5$ & $37.8\pm1.6$ & $38.3\pm2.5$ & $38.5\pm1.3$ & $39.5$\tabularnewline
{\small{}$7,5,3$} & $43.3\pm2.9$ & $38.7\pm2.4$ & $38.6\pm2.1$  & $38.8\pm0.9$ & $40.2\pm2.0$ & $37.9\pm1.7$ & $39.7\pm0.7$ & $39.6$\tabularnewline
{\small{}$7,5,5$} & $43.5\pm2.5$ & $40.2\pm0.9$ & $40.8\pm2.6$ & $38.4\pm0.9$ & $40.8\pm1.5$ & $40.0\pm0.4$ & $41.7\pm2.5$ & $40.8$\tabularnewline
{\small{}$7,7,5$} & $43.5\pm2.7$ & $41.6\pm3.0$ & $43.3\pm6.1$ & $40.5\pm2.9$ & $39.8\pm2.5$ & $47.3\pm2.5$ & $41.6\pm2.0$ & $42.5$\tabularnewline
\hline 
Mean & $44.3$ & $41.2$ & $41.1$ & $40.4$ & $40.8$ & $41.2$ & $41.6$ & \tabularnewline
\end{tabular}
\par\end{centering}

\protect\caption{\label{tab:num-filters-vs-filters-size}Detection quality (MR$\%$)
as the number of filters per layer (columns) and filter sizes per
layer (rows). CifarNet parameters are highlighted in italic. (MR:
log-average miss-rate on Caltech validation set).}
\end{table*}

\section{\label{sec:AlexNet-grid-search}Grid search for AlexNet}

Table \ref{tab:svm-parameter-tuning} presents the swipe of parameters
used to construct the ``Best parameters'' entries in table 8 of
the main paper. We vary the criterion to select negative samples and
the SVM regularization parameters. Defaults are parameters are $\mbox{IoU}<0.5$,
and $\mbox{C}=10^{-3}$.

Overall we notice that neither parameter is very sensitive ($1\sim2$
percent points fluctuations). When C is far from optimal large degradation
is observed ($10$ per cent points). As seen in table 8 of the main
paper the gap between default and tuned parameters is rather small
($1\sim2$ percent points).
\begin{table*}
\begin{centering}
\subfloat[layer fc7]{\begin{centering}
\begin{tabular}{c|ccccccccc}
\diaghead{\hskip5.6em}{neg\\overlap}{C} & $10^{-6}$ & $10^{-5.5}$ & $10^{-5}$ & $10^{-4.5}$ & $10^{-4}$ & $10^{-3.5}$ & $10^{-3}$ & $10^{-2.5}$ & $10^{-2}$\tabularnewline
\hline 
\hline 
0.3 & 36.01\% & 33.62\% & 32.30\% & 32.22\% & 32.04\% & 32.42\% & 32.24\% & 32.26\% & 32.40\%\tabularnewline
0.4 & 36.01\% & 33.72\% & 32.43\% & 32.09\% & 32.16\% & 32.33\% & 32.23\% & 32.30\% & 32.20\%\tabularnewline
0.5 & 36.07\% & 33.90\% & 32.51\% & 32.03\% & 32.18\% & 32.53\% & 32.20\% & 32.28\% & 33.15\%\tabularnewline
0.6 & 36.50\% & 33.96\% & 32.43\% & 32.19\% & 32.24\% & 32.45\% & 32.29\% & 33.06\% & 34.61\%\tabularnewline
0.7 & 36.55\% & 34.32\% & 32.36\% & 32.05\% & 32.15\% & 32.55\% & 32.82\% & 33.83\% & 36.13\%\tabularnewline
\end{tabular}
\par\end{centering}

}
\par\end{centering}

\begin{centering}
\subfloat[layer fc6]{\begin{centering}
\begin{tabular}{c|ccccccccc}
\diaghead{\hskip5.6em}{neg\\overlap}{C} & $10^{-6}$ & $10^{-5.5}$ & $10^{-5}$ & $10^{-4.5}$ & $10^{-4}$ & $10^{-3.5}$ & $10^{-3}$ & $10^{-2.5}$ & $10^{-2}$\tabularnewline
\hline 
\hline 
0.3 & 37.16\% & 32.49\% & 32.01\% & 31.88\% & 32.03\% & 32.18\% & 32.50\% & 32.40\% & 32.48\%\tabularnewline
0.4 & 37.16\% & 32.54\% & 32.07\% & 31.89\% & 32.14\% & 31.92\% & 32.46\% & 32.51\% & 32.56\%\tabularnewline
0.5 & 37.41\% & 32.61\% & 32.17\% & 32.07\% & 32.04\% & 31.84\% & 32.57\% & 33.12\% & 33.18\%\tabularnewline
0.6 & 37.54\% & 32.68\% & 32.14\% & 32.12\% & 32.22\% & 31.90\% & 32.93\% & 34.02\% & 35.85\%\tabularnewline
0.7 & 38.06\% & 32.67\% & 32.10\% & 31.89\% & 32.23\% & 32.32\% & 33.92\% & 35.92\% & 38.72\%\tabularnewline
\end{tabular}
\par\end{centering}

}
\par\end{centering}

\begin{centering}
\subfloat[layer pool5]{%
\begin{tabular}{c|ccccccccc}
\diaghead{\hskip5.6em}{neg\\overlap}{C} & $10^{-6}$ & $10^{-5.5}$ & $10^{-5}$ & $10^{-4.5}$ & $10^{-4}$ & $10^{-3.5}$ & $10^{-3}$ & $10^{-2.5}$ & $10^{-2}$\tabularnewline
\hline 
\hline 
0.3 & 55.37\% & 36.77\% & 33.16\% & 32.75\% & 32.77\% & 33.29\% & 33.37\% & 34.28\% & 35.16\%\tabularnewline
0.4 & 55.89\% & 36.82\% & 33.17\% & 32.52\% & 32.82\% & 33.16\% & 32.79\% & 34.12\% & 35.42\%\tabularnewline
0.5 & 56.24\% & 37.09\% & 33.21\% & 32.65\% & 32.69\% & 33.14\% & 33.26\% & 34.95\% & 36.39\%\tabularnewline
0.6 & 56.68\% & 37.19\% & 33.40\% & 32.66\% & 32.83\% & 33.44\% & 34.17\% & 35.66\% & 38.28\%\tabularnewline
0.7 & 57.93\% & 37.60\% & 33.81\% & 32.85\% & 33.27\% & 34.23\% & 35.76\% & 38.98\% & 42.68\%\tabularnewline
\end{tabular}

}
\par\end{centering}

\centering{}\subfloat[layer conv4]{%
\begin{tabular}{c|ccccccccc}
\diaghead{\hskip5.6em}{neg\\overlap}{C} & $10^{-6}$ & $10^{-5.5}$ & $10^{-5}$ & $10^{-4.5}$ & $10^{-4}$ & $10^{-3.5}$ & $10^{-3}$ & $10^{-2.5}$ & $10^{-2}$\tabularnewline
\hline 
\hline 
0.3 & 82.29\% & 64.90\% & 48.26\% & 44.67\% & 44.83\% & 43.66\% & 42.71\% & 43.36\% & 45.48\%\tabularnewline
0.4 & 82.29\% & 65.06\% & 48.66\% & 44.69\% & 44.67\% & 43.06\% & 42.41\% & 42.74\% & 44.81\%\tabularnewline
0.5 & 82.22\% & 65.23\% & 48.87\% & 44.68\% & 44.34\% & 42.98\% & 42.57\% & 43.30\% & 44.98\%\tabularnewline
0.6 & 82.22\% & 65.30\% & 48.69\% & 44.89\% & 44.39\% & 43.63\% & 42.92\% & 44.27\% & 46.35\%\tabularnewline
0.7 & 82.39\% & 65.96\% & 50.47\% & 45.62\% & 45.32\% & 44.86\% & 44.84\% & 46.31\% & 50.13\%\tabularnewline
\end{tabular}

}\protect\caption{\label{tab:svm-parameter-tuning}Detection quality (MR) as function
of the maximal IoU threshold to consider a proposal as negative example
and the SVM regularization parameter C. (MR: log-average miss-rate
on Caltech validation set)}
\end{table*}

\section{\label{sec:Datasets-statistics}Datasets statistics}

\begin{figure}
\begin{centering}
\subfloat[\label{fig:Caltech-sizes-histogram}Caltech Reasonable training set]{\begin{centering}
\includegraphics[width=1\columnwidth]{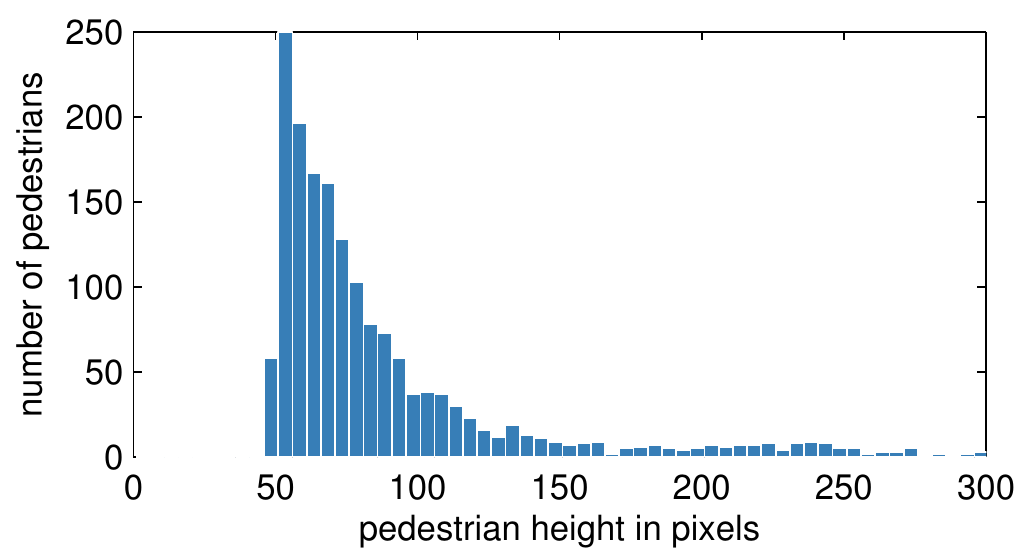}
\par\end{centering}

}
\par\end{centering}

\begin{centering}
\subfloat[\label{fig:KITTI-sizes-histogram}KITTI training set]{\begin{centering}
\includegraphics[width=1\columnwidth]{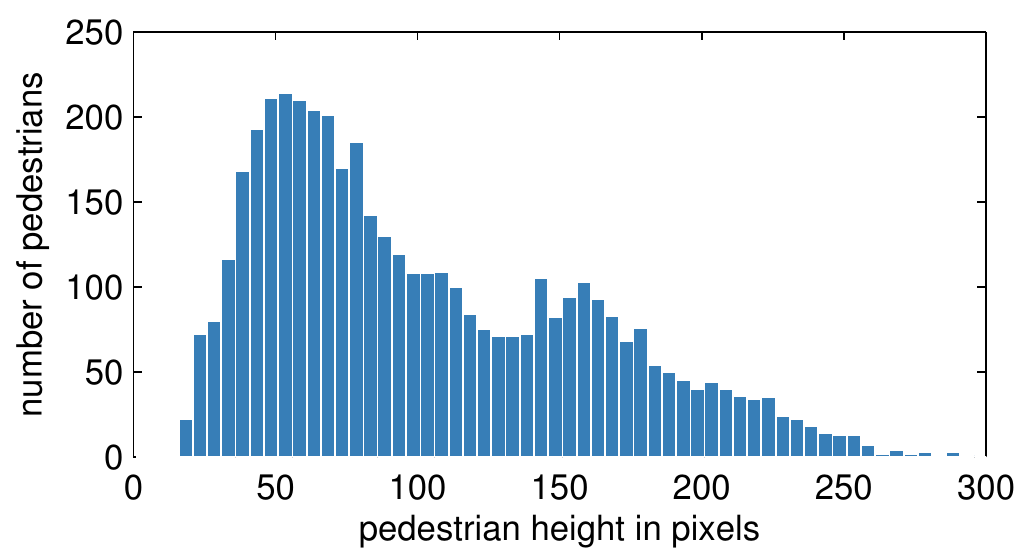}
\par\end{centering}

}
\par\end{centering}

\protect\caption{\label{fig:Pedestrian-size-histograms}Histogram of pedestrian heights
in different datasets.}
\end{figure}

In figure \ref{fig:Pedestrian-size-histograms} we plot the height
distribution for pedestrians in Caltech and KITTI training sets. Although
the datasets are visually similar, the height distributions are somewhat
dissimilar (for reference ImageNet and Pascal distributions are more
look alike among each other).

It was shown in \cite{Benenson2014Eccvw} that models trained in each
dataset, do not transfer well across each other (compared to models
trained on the smaller INRIA dataset).

\begin{figure}
\begin{centering}
\includegraphics[width=1\columnwidth]{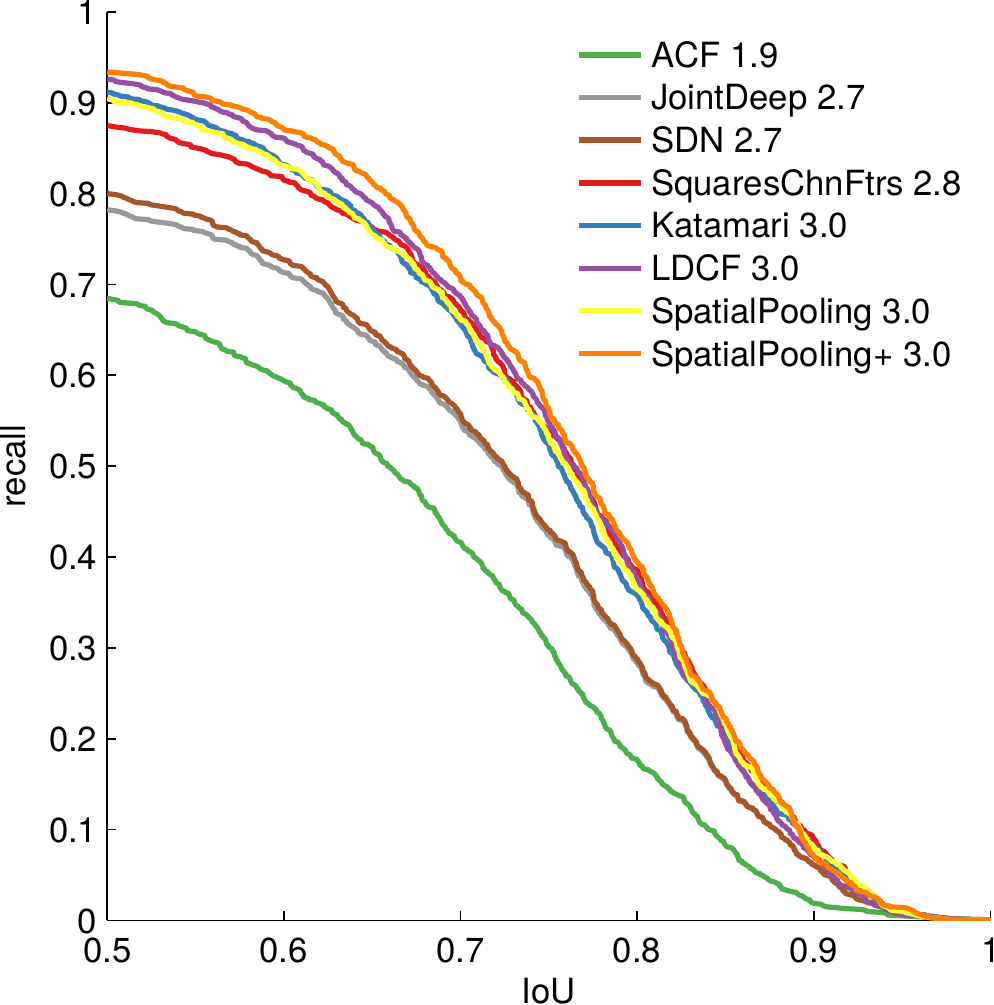}\protect\caption{\label{fig:Recall-vs-iou}Recall of ground truth versus IoU threshold,
for a selection of detection methods. The curves are cumulative distributions.
The detections have been filtered by score to reach $\sim\!\!3$ proposals
per image on average (number indicated in the legend).}

\par\end{centering}

\begin{centering}
\vspace{2em}
\includegraphics[width=1\columnwidth]{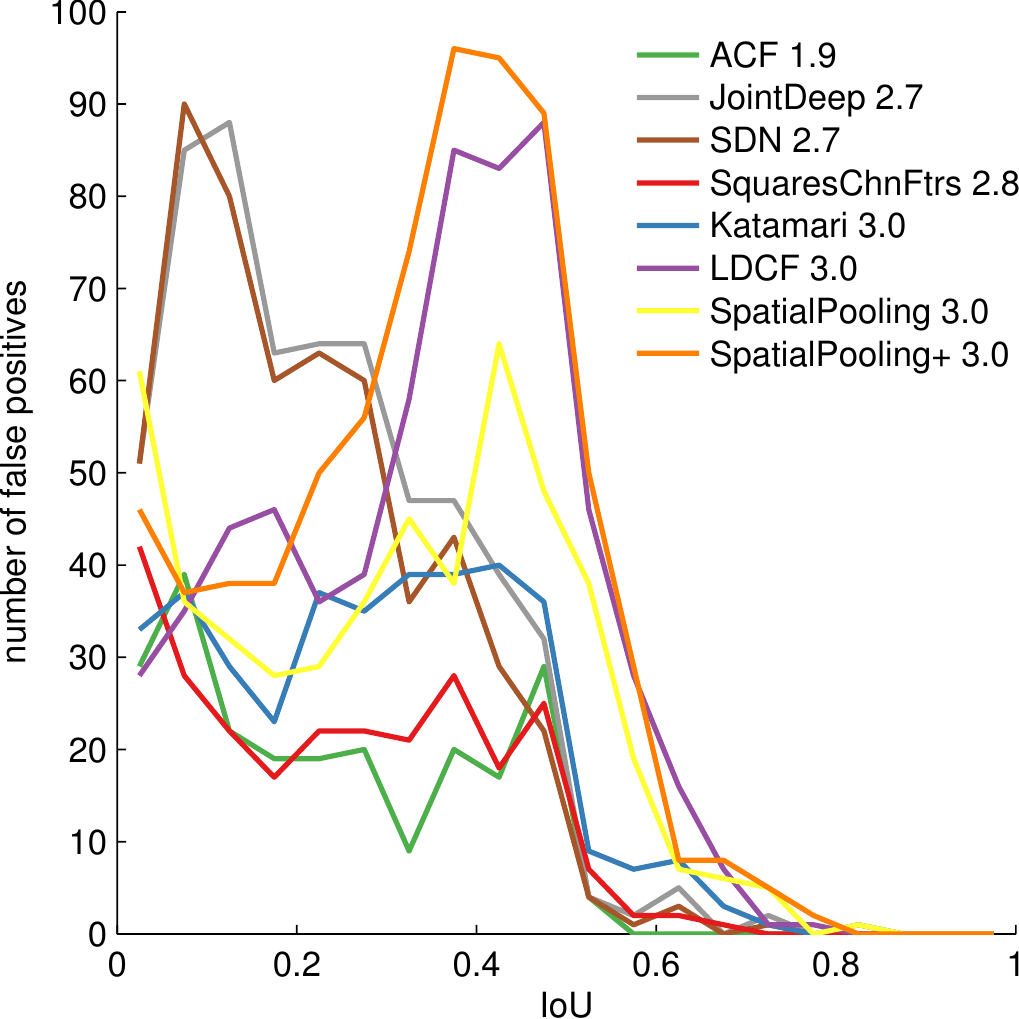}
\par\end{centering}

\protect\caption{\label{fig:fp-vs-iou-proposals}Distribution of overlap between false
positives and ground truth of those false positives that do overlap
with the ground truth. The curves are histogram with coarse IoU bins.
Number in the legend indicates the average number of proposals per
image (after filtering to reach $\sim\!\!3$).}
\end{figure}

\section{\label{sec:Proposal-statistics}Proposals statistics}

In figures \ref{fig:Recall-vs-iou} and \ref{fig:fp-vs-iou-proposals}
we show statistics of different detectors on the Caltech test set,
including the ones we use as proposals in our experiments. These figures
complement table 9 of the main paper.

Our initial experiments indicated that it is important to keep a low
number of average proposals per image in order to reduce the false
positives rate (post re-scoring). This is in contrast to common practice
when using class-agnostic proposal methods, where using more windows
is considered better because they provide higher recall \cite{Hosang2014Bmvc}.
We filter proposals via a threshold on the detection score.

As can be seen in figure \ref{fig:Recall-vs-iou} a recall higher
than $90\%$ can be achieved with only $\sim\!\!3$ proposals per
image on average (for Intersection-over-Union threshold above $0.5$,
the evaluation criterion). The average number of proposals per image
is quite low because most frames of the Caltech test set do not contain
any pedestrian. 

In figure \ref{fig:fp-vs-iou-proposals} we show the number of false
positives at different overlap levels with the ground truth annotations.
The bump around $0.5$ IoU, most visible for \texttt{Spatial\-Pooling}
and \texttt{LDCF}, is an artefact of the non-maximum suppression method
used by each method. Both these method obtain high quality detection,
thus they must assign (very) low-scores to these false positives windows.
To further improve quality the re-scoring method must do the same.

When using a method for proposals one desires to have high recall
with high overlap with the ground truth (figure \ref{fig:Recall-vs-iou}),
as well has having false positives with low overlap with the ground
truth (figure \ref{fig:fp-vs-iou-proposals}). False positive proposals
overlapping true pedestrians will have pieces of persons, which might
confuse the re-scoring classifier. Classifying fully centred persons
versus random background is assumed to be easier task.

In table 9 of the main paper we see that AlexNet reaches top detection
quality by improving over \texttt{LDCF}, \texttt{SquaresChnFtrs},
and \texttt{Katamari}. 
\end{document}